\documentclass[runningheads]{llncs}
\usepackage{graphicx}
\usepackage{comment}
\usepackage{amsmath,amssymb} 
\usepackage{color}

\usepackage{array,colortbl,multirow,multicol,booktabs,ctable}
\usepackage{float}  
\usepackage{subfigure}
\usepackage{caption}
\usepackage{amsfonts}
\usepackage{nicefrac}
\usepackage{microtype}
\usepackage{amssymb}
\usepackage{xcolor}
\usepackage{wrapfig,lipsum,booktabs}
\newcommand{\ie}{\textit{i}.\textit{e}.}

\newcommand\blue[1]{\textcolor{blue}{\textbf{#1}}}
\newcommand\red[1]{\textcolor{red}{\textbf{#1}}}
\newcommand\bluetext[1]{\textcolor{blue}{#1}}
\newcommand\redtext[1]{\textcolor{red}{#1}}
\newcommand\boldtext[1]{\textbf{#1}}

\makeatletter
\newcommand\figcaption{\def\@captype{figure}\caption} 
\newcommand\tabcaption{\def\@captype{table}\caption} 
\makeatother
\usepackage{xspace}
\makeatletter
\DeclareRobustCommand\onedot{\futurelet\@let@token\@onedot}
\def\@onedot{\ifx\@let@token.\else.\null\fi\xspace}
\def\eg{\emph{e.g}\onedot} 
\def\ie{\emph{i.e}\onedot} 
 
\def\etc{\emph{etc}\onedot}

\makeatother

\usepackage[ruled]{algorithm2e}

\begin{document}
\pagestyle{headings}
\mainmatter
\def\ECCVSubNumber{1257}  

\title{HAVANA: \\Hierarchical and Variation-Normalized Autoencoder for Person Re-identification} 


\titlerunning{Hierarchical and Variation-Normalized Autoencoder}
%
\author{
Jiawei Ren*\inst{1} \and
Xiao Ma*\inst{2} \and
Chen Xu*\inst{2} \and
Haiyu Zhao$^\dagger$\inst{1} \and
Shuai Yi\inst{1} 
}
\authorrunning{Ren J., Ma X., Xu C., Zhao H., Yi S.}
%
\institute{
SenseTime Research, \email{renjiawei@sensetime.com} \and
National University of Singapore, \email{xiao-ma@comp.nus.edu.sg}
\\}
\maketitle

\begin{abstract}
Person Re-Identification (Re-ID) is of great importance to the many video surveillance systems. Learning discriminative features for Re-ID remains a challenge due to the large variations in the image space, e.g., continuously changing human poses, illuminations and point of views. In this paper, we propose HAVANA, a novel extensible, light-weight HierArchical and VAriation-Normalized Autoencoder that learns features robust to intra-class variations. In contrast to existing generative approaches that prune the variations with heavy extra supervised signals, HAVANA suppresses the intra-class variations with a Variation-Normalized Autoencoder trained with no additional supervision. We also introduce a novel Jensen-Shannon triplet loss for contrastive distribution learning in Re-ID. In addition, we present Hierarchical Variation Distiller, a hierarchical VAE to factorize the latent representation and explicitly model the variations. 
To the best of our knowledge, HAVANA is the first VAE-based framework for person ReID.

\end{abstract}

\renewcommand{\thefootnote}{\fnsymbol{footnote}}
\footnotetext[1]{Equal contribution}
\footnotetext[4]{Corresponding author}
\renewcommand{\thefootnote}{\arabic{footnote}}

\section{Introduction}
Pedestrian Re-Identification (Re-ID) aims to associate pedestrian images across different cameras and time periods. An efficient robust Re-ID system is of fundamental importance to various surveillance applications. For example, finding a lost child in the mall, searching for a suspect at the airport and etc~\cite{zhao2017spindle}. 

As a well-studied problem, Re-ID, however, remains challenging and an unsolved task. One major challenge is the large variations in the image space. It is common that a camera does not always capture the human face and the Re-ID model has to be able to match the back view of a pedestrian with the front view. The constantly changing human poses, illumination conditions and point of views also requires a robust Re-ID model against the variations. 

Most state-of-the-art Re-ID models often aim to learn invariant features to be insensitive against the noise. In particular, with the recent advances of the deep generative models, GANs have been widely used to generate clean representations from the noisy observations. However, these models normally rely on additional supervised training signals, \eg, keypoints/poses, camera-styles, \etc, which not only requires heavy data labeling efforts, but also burdens the training process. 

In this paper, we propose HAVANA, a novel extensible, light-weight HierArchical and VAriation-Normalized Autoencoder framework for Re-ID with explicit variation modeling and no extra data labeling. In contrast to the prior works that focus on filtering the variations, HAVANA adopts a \emph{Variational Autoencoder} (VAEs)~\cite{kingma2013auto} based structure, that explicitly models the feature variation with a learned latent variable. HAVANA consists of three main components. First, we introduce \emph{Variation-Normalized Autoencoder} (VNAE) framework, which is the base of HAVANA, that applies standard supervised Re-ID training signals, classification loss and triplet loss, to regularize the variance of the latent distribution in VAE. Specifically, we backpropagate the supervised gradients to the mean of the latent distribution, ignoring the noisy variations, which automatically learns the shared clean feature of an identity. Second, we introduce a novel variance aware triplet loss function, \emph{Jensen-Shannon Triplet Loss} (JS triplet loss), that computes the contrastive loss of two distributions. JS triplet loss encourages the VNAE to consider distributional distance and improves the robustness of VNAE against the noise. Third, \emph{Hierarchical Variation Distiller} (HVD) is proposed to further improve variation filtering and disentangled representation learning. HVD factorizes the latent variations into a hierarchical structure with two additional variables to model high-level variations.


We experiment on three large scale datasets: Market-1501, DukeMTMC-reID and MSMT17. We train HAVANA by only the unsupervised ELBO loss, and the combination of Jensen-Shannon triplet loss and cross-entropy loss as a supervised signal for Re-ID, without any extra supervision. Our results significantly outperform the state-of-the-art (SOTA) generative approaches that extensively rely on other labels. Comparing with all existing Re-ID methods, we achieve SOTA results on two datasets and strong performance on the rest.

We summarize our contributions as follows: 1) we propose HAVANA, a VAE-based Re-ID framework with minimal data labeling requirements that achieves the SOTA performance on 3 commonly used Re-ID benchmarks; 2) we introduce Jensen-Shannon triplet loss, a robust variance aware contrastive feature learning loss function for Re-ID; 3) we design a hierarchical variation distiller module to for a factorized variation modeling. 

\section{Background}
\subsection{Variational Autoencoders}\label{sect:vae}
Variational Autoencoder (VAE) \cite{kingma2013auto} is an unsupervised representation learning algorithm that embeds high-dimensional observations into a low-dimensional space such that the learned embeddings, \ie, meta-prior \cite{BengioRepresentation}, can reconstruct to the original observations. A VAE has an encoder-decoder structure: we encode observation $\mathbf{x}$ into a latent variable $\mathbf{z}$ and decode into $\mathbf{x}$ again. Specifically, we define a family of prior distribution $p_\theta (\mathbf{z})$ over the latent variable $\mathbf{z}$ and decoder $p_\theta (\mathbf{x}\mid \mathbf{z})$ over observations $\mathbf{x}$. A VAE learns to maximize the observation likelihood $\sum_{n=1}^N \log p_\theta(\mathbf{x}^n)$, of a given dataset $\{\mathbf{x}^n\}_{n=1}^N$. The observation likelihood can further be factorized as $\int p_\theta(\mathbf{x}\mid \mathbf{z})p_{\theta}(\mathbf{z})dz$. VAE proposes to tackle the intractable log marginal likelihood problem by importance sampling with a encoder $q_\phi (\mathbf{z}\mid \mathbf{x})$, \ie, a conditional distribution of $\mathbf{z}$ given $\mathbf{x}$, and maximizes the Evidence Lower Bound (ELBO) of the observation likelihood, 
\begin{align}
    \mathrm{ELBO} &= \mathbb{E}_{q_\phi(\mathbf{z}\mid \mathbf{x})}\left[\log \frac{p_\theta (\mathbf{x}\mid \mathbf{z}) p_\theta(\mathbf{z})}{q_\phi (\mathbf{z}\mid \mathbf{x})}\right]\\
    &= \mathbb{E}_{q_\phi(\mathbf{z}\mid \mathbf{x})}\left[\log p_\theta (\mathbf{x}\mid \mathbf{z})\right] - D_{\mathrm{KL}}(q_\phi (\mathbf{z}\mid \mathbf{x}) \parallel p_\theta (\mathbf{z}))\label{eqn:vae}
\end{align}
where $D_\mathrm{KL}$ is the Kullback–Leibler divergence (KL Divergence).

The intuition behind VAE is that we sample $\mathbf{z}$ from encoder $q_\phi (\mathbf{z}\mid \mathbf{x})$ and reconstruct $x$ with decoder $p_\theta (\mathbf{x}\mid \mathbf{z})$. We want to minimize the distance between the encoder $q_\phi (\mathbf{z}\mid \mathbf{x})$ and the prior $p_\theta (\mathbf{z})$, as well as maximize the observation likelihood. In implementations, the distributions are assumed to be Gaussians and $p_\theta (\mathbf{z})$ is usually assumed as a diagonal Normal distribution $\mathcal{N}(\mathbf{0}, \mathbf{I})$ for simplicity.

\subsection{Related Works}
\subsubsection{Deep Person Re-ID}
Person Re-ID studies the person retrieval problem under a multi-camera setup \cite{Gong_2014_Springer}. Great progress has been achieved in the field lately 
\cite{Yu_2019_ICCV,He_2019_ICCV,Guo_2019_ICCV,Dai_2019_ICCV,Xia_2019_ICCV,Luo_2019_ICCV,Liu_2019_ICCV,Sun_2019_CVPR,Hou_2019_CVPR,Chen_2019_ICCV,Miao_2019_ICCV,Zhou_2019_ICCV,Quan_2019_ICCV,Yang_2019_CVPR,Alemu_2019_ICCV,Zhang_2019_CVPR,Zheng_2019_CVPR,zhao2017spindle}.
Many strong feature extractors has been designed, including deep neural networks that are attention guided, body-part aligned, keypoint aligned, semantic aligned, \etc. A recent survey \cite{Ye2020DeepLF} reviews the existing methods in detail. 


Among many approaches, multiple prior works attempt to utilise deep generative models to help improve the feature embedding. In particular, \cite{Zheng_2017_ICCV} initiated the trend by using GAN to generate unlabeled pseudo images to supplement limited image-per-identity in Re-ID datasets. Later, \cite{Liu_2018_CVPR} proposed a generative network to transfer pose variations from a large-scale dataset to target dataset for robust feaure learning. Similarly, \cite{Qian_2018_ECCV} fused the feature embeddings from the original image and a GAN-synthesized image with a different pose and thus normalised the human-poses. \cite{NIPS2018_7398} designed FD-GAN to distill a pose-unrelated feature by pruning the pose embedding learned by GAN. Meanwhile, \cite{Zhong2017CameraSA} employed CycleGAN to adapt labeled images to different camera styles and hence augment the training set. So far, all preceding methods treat the generative model as a standalone module. \cite{Zheng_2019_CVPR} for the first time proposed a pipeline to jointly train a shared encoder that is supervised by both discriminative loss and generative loss under the generative-adversarial framework. However, one major issue of GAN-based methods is that training GAN has been notoriously difficult and unstable \cite{NIPS2017_7159,Arjovsky2017WassersteinG}.



\subsubsection{Disentangled Representation Learning for VAEs}

\cite{Tschannen2018RecentAI} presents recent progresses in the field. 
The characteristics of the aforementioned meta-priors varies for downstream applications, for example, image-to-image translation\cite{NIPS2018_7404}, manipulating attributes on images\cite{NIPS2017_7178}. One group of meta-prior \cite{higgins2017beta,Kim2018DisentanglingBF,Chen2018IsolatingSO,Zhao2017InfoVAEIM,Louizos2015TheVF} encourages disentanglement, which factorizes variations in observations like viewpoint, lighting condition into independent dimensions of the meta-prior. In the same time, another group of works \cite{Gulrajani2016PixelVAEAL,Snderby2016LadderVA,Kingma2014SemisupervisedLW,Chen2016VariationalLA,Oord2017NeuralDR} proposed hierarchical VAEs that attempt to describe the observations with hierarchical attributes.

Our framework HAVANA is a combination of the above-mentioned approaches: we propose a hierarchical $\beta$-VAE approach that improves the disentangled latent representation learning. To the best of our knowledge, HAVANA is the first work to perform feature learning on the Person Re-ID task with VAEs. It also eases the training process compared with prior GAN-based generative approaches.

\section{Proposed Method}
\begin{figure}[t]
  \centering
  \includegraphics[width=1.0\linewidth]{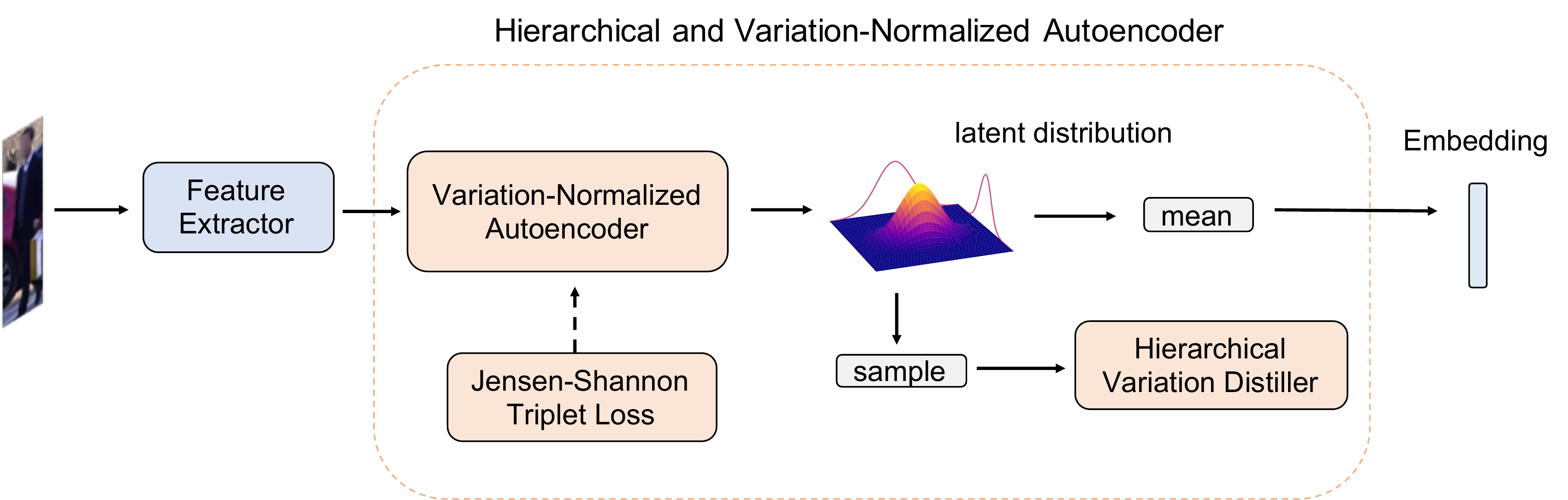}
  \caption{Overview of our proposed HAVANA framework.}
  \label{fig:overview}
  \vspace{-5mm}
\end{figure}
\subsection{Overview}
In this section, we present our VAE-based, joint generative and discriminative learning framework, Hierarchical and Variation-Normalized Autoencoder. The general Person Re-ID pipeline uses person identity labels to supervise a feature extractor that embeds the cropped person images into multi-dimensional feature vectors. Treating the feature vectors as observations, we propose a Variation-Normalized Autoencoder that normalizes the intra-class variations in the embedding space. To facilitate the above variational modeling process, we introduce a novel metric learning loss, Jensen-Shannon triplet loss, to assimilate the latent distributions of positive pairs and separate the latent distributions of negative pairs. Additionally, we distill the features by learning a structural representation of features with  Hierarchical Feature Distiller and meanwhile explicitly model the intra-class variation in an unsupervised manner. Lastly, we wrap up all proposed modules/losses in a holistic framework, and introduce a novel constraint to practically improve the framework's performance on multiple Person Re-ID datasets. Fig. \ref{fig:overview} demonstrates an overview of our method.

\subsection{Variation-Normalized Autoencoder for Re-identification}
\begin{figure}[t]
  \centering
  \includegraphics[width=1\linewidth]{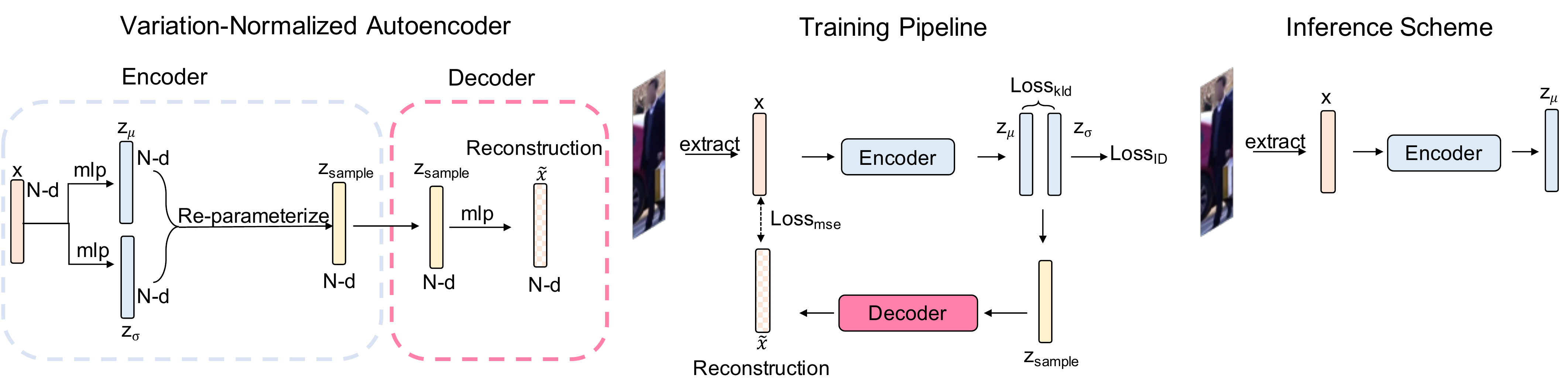}
  \caption{Network design, training/inference scheme of VNAE. VNAE consists of an encoder and a decoder. We first use a trained feature extractor to embedd an image to a feature vector $\textbf{x}$. Then $\textbf{x}$ is encoded to latent distribution $\textbf{z}$. $\textbf{z}$ is supervised by both ID loss and KL divergence. A sample of $\textbf{z}$ is decoded to $\tilde{\textbf{x}}$ to compute reconstruction loss with input $\textbf{x}$. During inference, VNAE directly outputs the encoded mean vector $\textbf{z}_\mu$ as the new embedding vector.}
  \label{fig:vnae}
  \vspace{-5mm}
\end{figure}
Prior works designed powerful feature extractors to learn discriminative features from person images with supervision of person identity labels. However, since the image-per-identity in most Re-ID datasets is limited \cite{Zheng_2017_ICCV}, the intra-class variations, e.g., poses and illumination changes, cannot be successfully filtered.
To alleviate the issue, we introduce Variation-Normalized Autoencoder (VNAE) to learn a variation-insensitive representation from the embedding features. 

We first employ a feature extracting network, which can be selected from any existing Person Re-ID models and not restricted to a specific type, to embed a cropped person image into a $N$-dimensional feature vector $\mathbf{x} \in \mathbb{R}^{N}$. We then build our VNAE over the feature $x$, using VAE described in Sect.~\ref{sect:vae}. Specifically, we first encode $\mathbf{x}$ with encoder $q_\phi(\mathbf{z}\mid \mathbf{x})$ and sample $\mathbf{z}$ from it:
\begin{gather}
    \mathbf{z}\sim q_\phi(\textbf{z}\mid \textbf{x}),\quad q_\phi(\textbf{z} \mid \textbf{x}) = \mathcal{N}(\textbf{z}_\mathbf{\mu}, \textbf{z}_\mathbf{\sigma}^2)\\
    \mathbf{z}_\mu = f_\mu (\mathbf{x}), \quad \mathbf{z}_\sigma = f_\sigma (\mathbf{x})
\end{gather}
where $\mathbf{z}_{\mu}$ and $\mathbf{z}_\sigma$ are the corresponding mean and standard deviation of the Gaussian distribution $q_\phi(\mathbf{z} \mid \mathbf{x})$, learned by two parameterized functions $f_\mu$ and $f_\sigma$ with $\mathbf{x}$ as input. Normally, $f_\mu$ and $f_\sigma$ are simple MLPs with non-linear activations. To achieve differentiable sampling, we apply the reparameterization trick~\cite{kingma2013auto}: we first sample $\epsilon \sim \mathcal{N}(\textbf{0},\textbf{I})$ and set $z=\textbf{z}_\mathbf{\mu} + \epsilon*\textbf{z}_\mathbf{\sigma}$. Lastly, we decode to observation $\mathbf{x}$ with decoder $p_\theta (\mathbf{x}\mid \mathbf{z})$. Fig. \ref{fig:vnae} shows VNAE in detail.

VNAE differs from VAE in terms of the variation normalization during training and inference. Specifically, Re-ID task requires distinctive features for different identities. 
To disentangle the identities, we supervise $\mathbf{z}_\mu$ with a cross-entropy Re-ID classification loss and triplet loss, where the classification loss learns to predict the correct identity of $\mathbf{x}$ using $\mathbf{z}_\mu$ , and the triplet loss learns to separate the embeddings $\mathbf{z}$ of different identities in the latent space. The gradient optimizes the mean of the random variable $\mathbf{z}$ to represent the common features of an identity and filters the variance of the distribution, thus regularizes the variations. A detailed definition can be found in the following sections. As a result, VNAE is optimized by 
\begin{align}
    \mathcal{L}_{\mathrm{VNAE}} &= \mathcal{L}_\mathrm{cls} + \lambda\mathcal{L}_\mathrm{triplet} - \alpha\mathrm{ELBO}_\mathrm{VNAE}\\
    \mathrm{ELBO}_\mathrm{VNAE} &= \mathbb{E}_{q_\phi(\mathbf{z}\mid \mathbf{x})}\left[\log p_\theta (\mathbf{x}\mid \mathbf{z})\right] -\beta D_{\mathrm{KL}}(q_\phi (\mathbf{z}\mid \mathbf{x}) \parallel p_\theta (\mathbf{z}))
\end{align}
where $\lambda$ and $\alpha$ are hyper-parameters for balancing the losses, and $\mathrm{ELBO}_\mathrm{VNAE}$ follows the $\beta$-VAE~\cite{higgins2017beta} setup that encourages to learn disentangled representation with $\beta$ parameter. Together with the discriminative losses $\mathcal{L}_\mathrm{triplet}$ and $\mathcal{L}_\mathrm{cls}$, VNAE would be able to learn a disentangled latent representation, \ie, distinctive features, for Re-ID.

Comparing with methods that utilise GAN to generate pseudo-images 
with additional annotations, and thus augment the training set to remove a specific variation from the feature embedding, VNAE has following merits:
    1) VNAE models variation in an unsupervised manner and requires no additional annotations;
    2) VNAE is cost-efficient with a unified generative and discriminative model;
    3) VNAE is decoupled with the feature extraction methods and hence extensible to work with SOTA feature extractors.


\subsection{Jensen-Shannon Triplet Loss}


Triplet loss is a contrastive loss that optimizes toward letting the distance within positive pairs greater than the distance within negative pairs by a margin. Formally, for an anchor sample $z_\textrm{i}$, same-identity sample $z_\textrm{j}$ and different-identity sample $z_\textrm{k}$, the triplet loss  $\mathcal{L}_\mathrm{triplet}$ is defined as:
\begin{equation}
    \mathcal{L}_\mathrm{triplet} = \max(d(z_\textrm{i}, z_\textrm{j}) - d(z_\textrm{i}, z_\textrm{k}) + \gamma, 0)
\end{equation}
where $d(\cdot, \cdot)$ denotes Euclidean distance for a pair of samples and $\gamma$ is a pre-defined margin.
When applying to distribution means, the triplet loss can pull the centers of positive distribution pairs together and push the centers of negative distributions pairs away. However, conventional triplet loss ignores the distribution assumption of embeddings and might be susceptible to variations.

To alleviate the issue, we propose a novel metric learning loss for distribution learning, Jensen-Shannon triplet loss (JS triplet loss).
We measure the pairwise distance with the Jensen-Shannon divergence (JS divergence)
\begin{equation}
    D_{\textrm{JS}}(\textbf{p} \parallel \textbf{q})={\frac  {1}{2}}D_\mathrm{KL}(\textbf{p} \parallel \textbf{m})+{\frac{1}{2}}D_\mathrm{KL}(\textbf{q} \parallel \textbf{m})
\end{equation}
where $\textbf{m}={\frac  {1}{2}}(\textbf{p}+\textbf{q})$.

The JS divergence can adequately reflect the similarity between two distributions. We compute the JS triplet loss following the scheme of computing the triplet loss between two embedding vectors. For an anchor distribution $\textbf{z}_\textrm{i}$, a same-identity distribution $\textbf{z}_\textrm{j}$ and a different-identity distribution $\textbf{z}_\textrm{k}$, the JS triplet loss is:
\begin{equation}
     \mathcal{L}_\mathrm{jst} = \max (D_{\textrm{JS}}(\textbf{z}_\textrm{i}\parallel\textbf{z}_\textrm{j}) -D_{\textrm{JS}}(\textbf{z}_\textrm{i}\parallel\textbf{z}_\textrm{k})  + \gamma, 0)
\end{equation}
where 
$\gamma$ is the margin parameter.

JS triplet loss has two major advantages: 1) JS triplet loss is variance aware. It considers the sample variance when computing the loss, which identifies noisy samples with large variance and output a smaller loss that improves the robustness of the network; 2) as JS divergence is symmetric and bounded in [0, 1], $\mathcal{L}_\mathrm{jst}$ is bounded in $[0, 1+\gamma]$. This gives a robust loss function against outliers and a numerically stable training process.


\subsection{Hierarchical Variation Distiller}
\begin{figure}[t]
  \centering
  \includegraphics[width=1.0\linewidth]{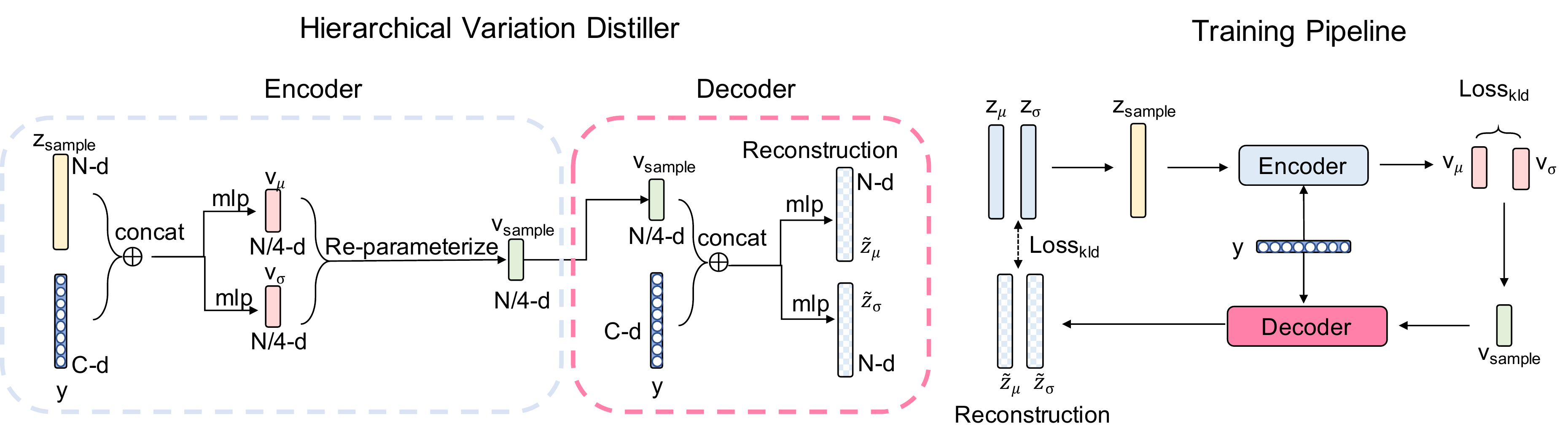}
  \caption{Network design and training scheme of HVD. HVD is of encoder-decoder structure. A sample of latent distribution $\textbf{z}$ is concatenated with the one-hot label $\textbf{y}$ to encode variation distribution $\textbf{v}$. A sample from $\textbf{v}$ is again concatenated with the one-hot vector $\textbf{y}$ to decode a reconstructed distribution $\tilde{\textbf{z}}$. Two unsupervised loss are computed.}
  \label{fig:hvd_arc}
  \vspace{-5mm}
\end{figure}
\begin{wrapfigure}{r}{0.25\textwidth}
\vspace{-10mm}
  \begin{center}
    \includegraphics[width=0.5\linewidth]{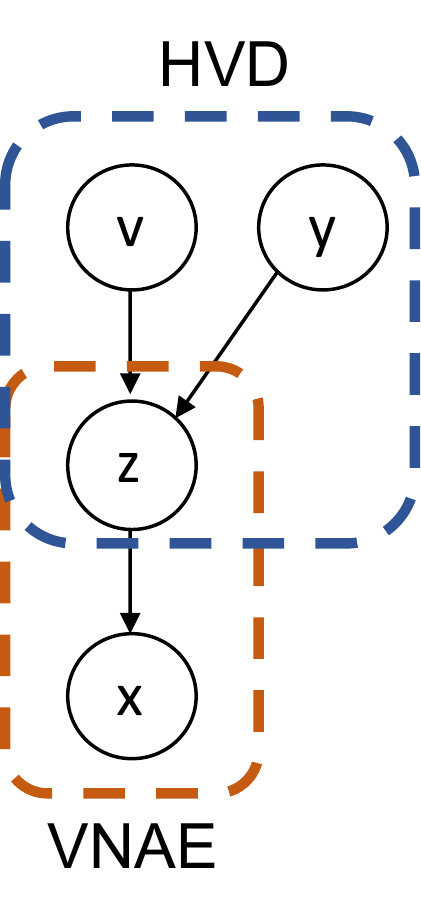}
  \end{center}
  \vspace{-5mm}
  \caption{\small Hierarchical variation distillation. HVD distills the variance of embedding variable $\mathbf{z}$ of observation $\mathbf{x}$ with variation variable $\mathbf{v}$ and identity variable $\mathbf{y}$.}
  \label{fig:hvd}
 \vspace{-4mm}
\end{wrapfigure}
We introduce the Hierarchical Variation Distiller (HVD) module that further improves VNAE variation modeling with a hierarchical conditional VAE.

HVD is a higher level variational autoencoder following the VNAE framework. Besides random variable $\mathbf{x}$ and $\mathbf{z}$ in VNAE, we introduce two additional random variables, variation variable $\mathbf{v}$ and identity variable $\mathbf{y}$. We model a joint distribution of $p(\mathbf{x}, \mathbf{z}, \mathbf{v}, \mathbf{y})$, factorized as
\begin{equation}
    p(\mathbf{x}, \mathbf{z}, \mathbf{v}, \mathbf{y}) = p(\mathbf{x}\mid \mathbf{z}) p(\mathbf{z}\mid \mathbf{v}, \mathbf{y}) p(\mathbf{v}) p(\mathbf{y})
\end{equation}
where its graphical model representation is given by Fig.~\ref{fig:hvd}. The sub-model, $p(\mathbf{z}\mid \mathbf{v}, \mathbf{y}) p(\mathbf{v}) p(\mathbf{y})$, forms the HVD part.
The intuition behind is: we hope that a higher level random variable $\mathbf{v}$ is able to capture the variations over the embedding of the same identity, e.g., pose variations and occlusions. An identity variable $\mathbf{y}$ is assumed to be independent of $\mathbf{v}$, \ie, variation should be universally applied to all identities. Specifically, we assume variation variable has a prior of $p_\theta(\mathbf{v}) = \mathcal{N}(\mathbf{0, I})$ and identity variable follows a categorical distribution corresponding to the one-hot identity label $\mathbf{y}\sim \mathrm{Cat}(\mathbf{y})$.

To efficiently sample $\mathbf{v}$, we follow the VAE setup and sample $\mathbf{v}$ from a proposal distribution, \ie, encoder, $q_\phi(\mathbf{v}\mid \mathbf{z}, \mathbf{y})$. The encoder generates variation variable samples with higher likelihood to produce samples $\mathbf{z}$ and $\mathbf{y}$. As a result, this leads to a new ELBO for unsupervised learning:
\begin{equation}
\begin{aligned}
    \mathrm{ELBO}_\mathrm{HVD} = &\log \mathbb{E}_{q_\phi (\mathbf{z}\mid \mathbf{x}) q_\phi (\mathbf{v}\mid \mathbf{z}, \mathbf{y})}\left[ p_\theta(\mathbf{x}\mid \mathbf{z}) \right] + \log p_\theta(\mathbf{y})\\ &- D_\mathrm{KL}(q_\phi(\mathbf{z}\mid \mathbf{x}) \parallel p_\theta(\mathbf{z}\mid \mathbf{v}, \mathbf{y})) - D_\mathrm{KL}(q_\phi(\mathbf{v}\mid \mathbf{z}, \mathbf{y}) \parallel p_\theta(\mathbf{v})) \\
    & \propto \log \mathbb{E}_{q_\phi (\mathbf{z}\mid \mathbf{x}) q_\phi (\mathbf{v}\mid \mathbf{z}, \mathbf{y})}\left[ p_\theta(\mathbf{x}\mid \mathbf{z}) \right]\\ &- D_\mathrm{KL}(q_\phi(\mathbf{z}\mid \mathbf{x}) \parallel p_\theta(\mathbf{z}\mid \mathbf{v}, \mathbf{y})) - D_\mathrm{KL}(q_\phi(\mathbf{v}\mid \mathbf{z}, \mathbf{y}) \parallel p_\theta(\mathbf{v}))
\end{aligned}
\end{equation}
where $\log p_\theta(\mathbf{y})$ is a constant and can be ignored during optimization. 
The condition $\mathbf{y}$ encourages the network to learn a disentangled representation for each identity. The derivation of the $\mathrm{ELBO}_\mathrm{HVD}$ can be found in the supplementary material.

Essentially, HVD is an auxiliary module to the VNAE. We illustrate details of HVD in Fig. \ref{fig:hvd}. HVD improves the disentangled representation learning during training. It can be removed during inference and the remaining VNAE network keeps to be a light-weight Re-ID module.

\vspace{-5mm}
\begin{algorithm}[!htb]
\SetAlgoLined
\DontPrintSemicolon
\textbf{Input:} $\mathbf{x}_\textrm{img}$: cropped person image,
 $\mathbf{y}$: person identity label,
$\mathbf{extractor}(\cdot)$: trained feature extractor network\;
\While{not converged}{
    
    $\mathbf{x}\gets \textbf{extractor}(\mathbf{x}_\textrm{img})$\hfill (extract feature $\mathbf{x}$ from $\mathbf{x}_\textrm{img}$)
    
    $\mathbf{z_{\mu}}, \mathbf{z_{\sigma}} \gets \textrm{VNAE}\textrm{.encode}(\mathbf{x})$\hfill (map $\mathbf{x}$ to latent distribution $\mathbf{z}$)
    
    $\mathbf{z_\textrm{sample}} \gets \textrm{reparameterize}(\mathbf{z_{\mu}}, \mathbf{z_{\sigma}})$\hfill (sample from $\mathbf{z}$)
    
    $\mathbf{\tilde{x}} \gets \textrm{VNAE}\textrm{.decode}(\mathbf{z_\textrm{sample}})$\hfill (reconstruct to $\mathbf{x}$)
    
    $\mathbf{v_{\mu}}, \mathbf{v_{\sigma}} \gets \textrm{HVD}\textrm{.encode}(\mathbf{z_\textrm{sample}}, \mathbf{y})$\hfill (map $\mathbf{z}$ to variation distribution $\mathbf{v}$)
    
    $\mathbf{v_\textrm{sample}} \gets \textrm{reparameterize}(\mathbf{v_{\mu}}, \mathbf{v_{\sigma}})$\hfill (sample from $\mathbf{v}$)
    
    $\mathbf{\tilde{z}_{\mu}}, \mathbf{\tilde{z}_{\sigma}} \gets \textrm{HVD.decode}(\mathbf{v_\textrm{sample}}, \mathbf{y})$\hfill (reconstruct to distribution $\mathbf{z}$)
    
    $\mathcal{L}_\textrm{cls} \gets \textrm{CrossEntropy}
    (\textrm{classifier}(\mathbf{z}_\mathrm{\mu}), \mathbf{y})
    $
    
    $\mathcal{L}_\textrm{jst}\gets \textrm{JSTripletLoss}(\mathcal{N}(\mathbf{z_{\mu}}, \mathbf{z^\textrm{2}_{\sigma}}), \mathbf{y})$
    
    $\mathcal{L}_\textrm{reconx}\gets \textrm{MeanSquareError}(\mathbf{x},\mathbf{\tilde{x}}$) 
    
    $\mathcal{L}_\textrm{klz}\gets -D_\mathrm{KL}(\mathcal{N}(\mathbf{z_{\mu}}, \mathbf{z^\textrm{2}_{\sigma}}) \parallel N(\mathbf{\tilde{z}_{\mu}}, \mathbf{\tilde{z}^\textrm{2}_{\sigma}})$)
    
    $\mathcal{L}_\textrm{klv}\gets -D_\mathrm{KL}(\mathcal{N}(\mathbf{\tilde{v}_{\mu}}, \mathbf{v^\textrm{2}_{\sigma}}) \parallel \mathcal{N}(\textbf{0},\textbf{I}))$

    $\mathcal{L}_\textrm{total}\gets \mathcal{L}_\textrm{cls}  + \mathcal{L}_\textrm{jst} + \alpha (\mathcal{L}_\textrm{reconx}+\beta(\mathcal{L}_\textrm{klz} + \mathcal{L}_\textrm{klv})) $
    
    Backpropagate($\mathcal{L}_\textrm{total}$)
    
}
\textbf{Output:} $\mathbf{z}_{\mu}$: the output embedding vector 
\caption{HAVANA}
\label{algo:pseudocode}
\end{algorithm}
\vspace{-12mm}
\subsection{HAVANA}
Combining the two network modules, VNAE and  HVD, and one new metric learning loss, JS triplet loss, we present the holistic framework as Hierarchical and Variation-Normalized Autoencoder (HAVANA). HAVANA is a hybrid system with both unsupervised and supervised training signals. The overall loss is
\begin{align}
    \mathcal{L}_\mathrm{HAVANA} &= \mathcal{L}_\mathrm{cls} + \lambda \mathcal{L}_\mathrm{jst} - \alpha \mathrm{ELBO}_\textrm{HVD} \\
    \mathrm{ELBO}_\textrm{HVD} &=\log \mathbb{E}_{q_\phi (\mathbf{z}\mid \mathbf{x}) q_\phi (\mathbf{v}\mid \mathbf{z}, \mathbf{y})}\left[ p_\theta(\mathbf{x}\mid \mathbf{z}) \right]\nonumber\\ &-\beta \left(D_\mathrm{KL}(q_\phi(\mathbf{z}\mid \mathbf{x}) \parallel p_\theta(\mathbf{z}\mid \mathbf{v}, \mathbf{y})) + D_\mathrm{KL}(q_\phi(\mathbf{v}\mid \mathbf{z}, \mathbf{y}) \parallel p_\theta(\mathbf{v}))\right) \\
    & = \mathcal{L}_\mathrm{reconx} + \beta (\mathcal{L}_\mathrm{klz} + \mathcal{L}_\mathrm{klv})
\end{align}
where $\lambda$,  $\alpha$ and $\beta$ are hyper-parameters for loss ratio.

HAVANA benefits from the $\mathrm{ELBO}_\mathrm{HVD}$ for unsupervised disentangled representation learning with hierarchical variance regularization; simple extra supervised signals, including classification loss $\mathcal{L}_\mathrm{cls}$ and powerful Jensen-Shannon triplet loss $\mathcal{L}_\mathrm{jst}$ help to optimize the latent representation to generate robust and accurate Re-ID results. We summarize the overall pipeline in Alg.~\ref{algo:pseudocode}. 

In practice, we found that imposing a covariance constraint to variable $\mathbf{z}$ can significantly improve the performance. Specifically,  we enforce $\mathbf{z}_\sigma = \mathbf{I}$. The intuition is that the limited images-per-identity \cite{Zheng_2017_ICCV} might form a biased empirical distribution; covariance constraint softens the biased distribution and tries to recover the invariant identity distribution. We empirically validate the influence of covariance constraint in our ablation studies.

\begin{figure}[t]
  \centering
  \includegraphics[width=1.0\linewidth]{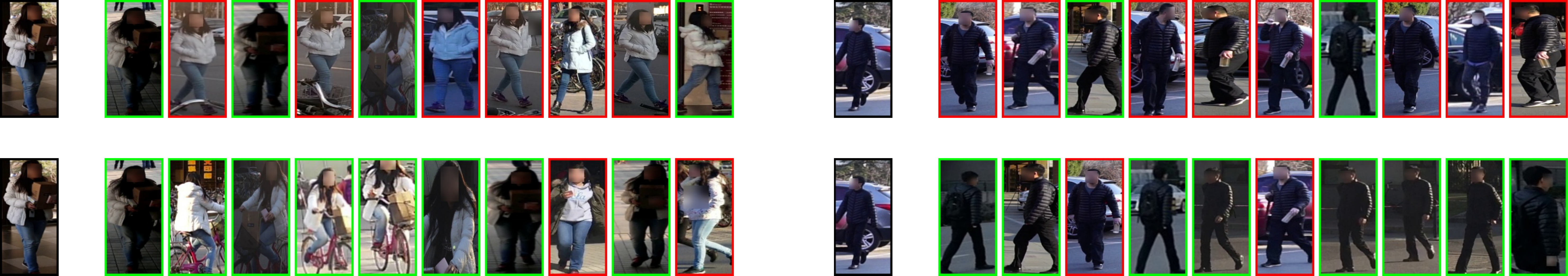}
  \rule[-.5cm]{4cm}{0cm}
  \vspace{-5mm}
  \caption{Comparison of top-10 retrieval results. Upper row: Baseline results. Lower row: HAVANA results. Images in green boxes are positive results and images in red boxes are negative results.}
  \vspace{-5mm}
  \label{fig:rank10}
\end{figure}

\begin{figure}[t]
  \centering
  \includegraphics[width=1\linewidth]{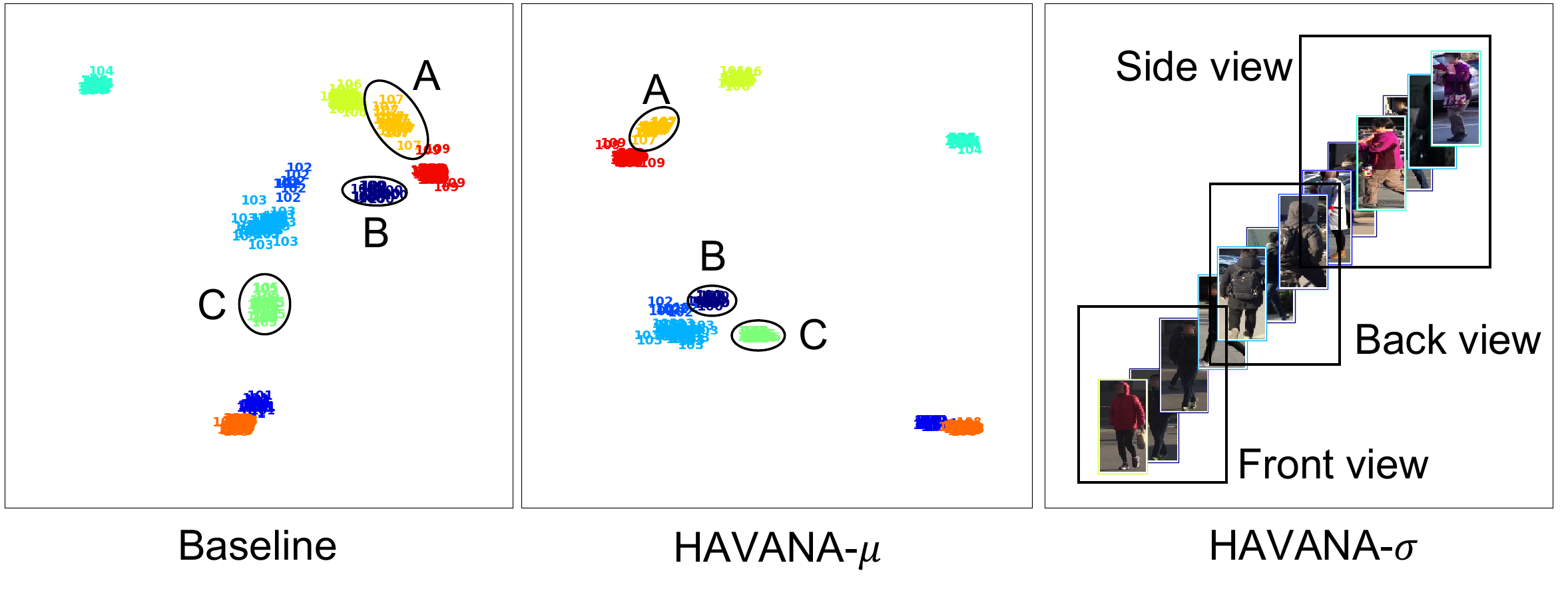}
  \rule[-.5cm]{4cm}{0cm}
  \vspace{-5mm}
  \caption{t-SNE visualization results. 10 random identities are sampled and represented in different colors. Baseline is the visualization of baseline embedding. HAVANA-$\mu$ is the visualization of HAVANA embedding, \ie, $\mathbf{z}_{\mu}$. HAVANA-$\sigma$ is the visualization of the standard deviation vector of the learned distribution, \ie, $\mathbf{z}_{\sigma}$. Regions A, B and C contains the embedding from same IDs across methods. Comparing with baseline, HAVANA achieves a more compact identity group in the embedding space. In HAVANA-$\sigma$, we observe that pose variation has been unsupervisingly clustered in the $\mathbf{z}_{\sigma}$ vector, which strongly advocates that HAVANA is capable of variation modeling.}
  \vspace{-5mm}
  \label{fig:tsne}
\end{figure}

\section{Experiments}

In this section, we evaluate our approach on three popular large-scale datasets: Market-1501, DukeMTMC-reID and MSMT17. We also perform extensive ablation studies to understand the influence of each proposed component.


\subsection{Experiment Setup and Implementation Details}
Evaluation is performed on Market-1501\cite{Zheng2015ScalablePR}, DukeMTMC-reID \cite{ergys2016performance} and MSMT17 \cite{Wei_2018_CVPR}. The detailed dataset statistics can be found in Table~\ref{tab:dataset}. Cumulative matching characteristics (CMC) Rank-1 accuracy and mAP are used as evaluation metrics.
We adopt ResNet-50 as our backbone. Our implementation is based on an open-source code-base \cite{Luo_2019_CVPR_Workshops,Luo_2019_Strong_TMM}. We follow their training schemes: the last stride of ResNet-50 is set from 2 to 1; the training and testing images are cropped into a size of 256x128; the dimension of output embedding is 2048; 
\begin{wraptable}{r}{0.5\linewidth}
\centering
\fontsize{7}{8}\selectfont
\begin{tabular}[b]{lccc}
        \toprule
        Dataset & \# IDs (T-Q-G) & \# images (T-Q-G)\\
        \hline\hline
        Market-1501 & 751-750-751 & 12936-3368-15913 \\
        DukeMTMC & 702-702-1110 & 16522-2228-17661 \\
        MSMT17 & 1041-3060-3060 & 30248-11659-82161  \\
        \bottomrule
        \end{tabular} 
        \caption{Dataset statistics. T: Train. Q: Query. G: Gallery.}
        \label{tab:dataset}
        \vspace{-4mm}
\end{wraptable}
learning rate warm-up, label smoothing \cite{Szegedy2015RethinkingTI}, random flip, Random Erasing \cite{Zhong2017RandomED} and BNNeck are employed; batch size is set to 64 with 4 images of 16 identities sampled each batch. We set the learning rate and weight decay to 1e-5 and 5e-4 respectively, and train the network for 300 epochs without learning rate decay. We have no test-time augmentation unless specified. We take ResNet-50 with the cross-entropy loss and the triplet loss as baseline, and train the baseline under above described setup. 
\begin{table}[!htb]

\centering
\begin{tabular}{ ll cc cc cc}
\toprule
Method & \multicolumn{2}{c}{Market-1501} &  \multicolumn{2}{c}{DukeMTMC} & \multicolumn{2}{c}{MSMT17} \\ \cmidrule{2-7}
       & \makebox[1cm][c]{mAP} & \makebox[1cm][c]{Rank-1} & \makebox[1cm][c]{mAP} & \makebox[1cm][c]{Rank-1} & \makebox[1cm][c]{mAP} & \makebox[1cm][c]{Rank-1} \\ \midrule
MGN~\cite{Wang2018LearningDF} (2018 ACMM) & 86.9 & \blue{95.7} & 78.4 & 88.7 & - & - \\
BoT \cite{Luo_2019_CVPR_Workshops} (2019 CVPRW ) & 85.9 & 94.5 & 76.4 & 86.4 & - & - \\
MHN~\cite{Chen_2019_ICCV} (2019 ICCV) & 85.0 & 95.1 & 77.2 & 89.1 & - & - \\
BDB~\cite{Dai_2019_ICCV} (2019 ICCV) & 86.7 & 95.3 & 76.6 & 89.0 & - & - \\
$P^2$-Net~\cite{Guo_2019_ICCV} (2019 ICCV) & 85.6 & 95.2 & 73.1 & 86.5 & - & - \\
SFT~\cite{Luo_2019_ICCV}(2019 ICCV) & 82.7 & 93.4 & 73.2 & 86.9 & 47.6 & 73.6\\
CAR~\cite{Zhou_2019_ICCV_Discriminative}(2019 ICCV) & 84.7 & 96.1  & 73.1 & 86.3 & - & - \\
BAT-net~\cite{Fang_2019_ICCV} (2019 ICCV) & 87.4 & 95.1 & 77.3 & 87.7 & 56.8 & 79.5 \\
SONA~\cite{Xia_2019_ICCV} (2019 ICCV) & \blue{88.8} & 95.6 & 78.3 & \blue{89.4} & - & -\\
ABD-Net~\cite{Chen_2019_ICCV_ABD-Net} (2019 ICCV) & 88.3 & 95.6 & \blue{78.6} & 89.0 & \blue{60.8} & \blue{82.30} \\
\midrule
Generative Approaches & & & & & & \\
\hline
USG-GAN\cite{Zheng_2017_ICCV} (2017 ICCV) & 56.2 & 78.1 & - & -&-&- \\
CSA-GAN\cite{Zhong2017CameraSA} (2018 CVPR) & 71.6& 89.5 &  57.6 & 78.3  &-&-\\
PN-GAN \cite{Qian_2018_ECCV}(2018 ECCV) & 72.6 & 89.4 & 53.2 & 73.6 & - & -\\
FD-GAN \cite{NIPS2018_7398} (2018 NIPS) & 77.7 & 90.5 & 64.5 & 80.0 & - & -\\
DGNet \cite{Zheng_2019_CVPR} (2019 CVPR)& 86.0 & 94.8 & 74.8 & 86.6 & 52.3 & 77.2 \\
\midrule
HAVANA & 86.9  & 94.4 & 77.4 & 87.7 & 52.3 & 76.8 \\
$\textrm{HAVANA}^{\dagger}$ & 87.4 & 94.5 & 77.9 &87.9 & 53.3 & 76.9 \\
$\textrm{HAVANA(MGN)}$ & 88.9 & 95.2 & 80.4 & 89.1 & 55.3 & 77.5 \\
$\textrm{HAVANA(MGN)}^{\dagger}$ & \red{89.2} & \red{95.3} & \red{80.8} & \red{89.4} & \red{56.0} & \red{77.9} \\
\bottomrule
\end{tabular}
\vspace{5mm}
\caption{Experimental comparison with state-of-the-art.  \redtext{Red} denotes our performance and \bluetext{blue} denotes the best performance achieved by existing methods. $\dagger$ denotes that the result is obtained with flip-test.}
\label{tab:sota}
\vspace{-10mm}
\end{table}
\subsection{Experiment Results}

We compare our model with state-of-the-art methods on Person Re-ID. We report our results on all three aforementioned datasets in Table~\ref{tab:sota}. Specifically, we compare HAVANA with the SOTA methods, including both non-generative Re-ID approaches and generative approaches.
With a baseline model with ResNet-50 backbone as the feature extractor, we show that our approach clearly outperforms all prior generative model based Person Re-ID methods. 
Replacing our feature extractor from the baseline model to a Multiple Granularities Network (MGN) \cite{Wang2018LearningDF} model  with ResNet-50 backbone, HAVANA achieves state-of-the-art results on both Market-1501 dataset and DukeMTMC dataset, meanwhile a strong result on MSMT17 dataset. The results concretely validate HAVANA's extensibility to be combined with other SOTA methods. Qualitative results are shown in Fig. \ref{fig:rank10} and Fig. \ref{fig:tsne}.

\subsection{Component Analysis}

\begin{table}[t]
    \begin{center}
      \fontsize{7}{8}\selectfont 
    \begin{tabular*}{0.8\textwidth}{l@{\extracolsep{\fill}}ccccccccc}
      \toprule
    V & J & H & C & Dataset &  mAP &  Rank-1 & Rank-5 & Rank-10\\
    \hline\hline
      & & & & MSMT17 & 48.2 & 72.2 & 83.9 & 87.7 \\
     \checkmark & & & & MSMT17 & 50.1 (+1.9) & 74.3 & 85.9 & 89.6 \\
     \checkmark & \checkmark & & & MSMT17 & 50.6 (+2.4) &  75.0 & 86.3 & 89.8 \\
     \checkmark & \checkmark & \checkmark & & MSMT17 & 52.0 (+3.8) & 75.9 & 87.2 & 90.3\\
     \checkmark & \checkmark & \checkmark & \checkmark & MSMT17 & 52.3 (+4.1)& 76.7 & 87.2 & 90.6\\

    \midrule
      & & & & Market-1501 & 85.8 & 94.2 & 98.4 & 99.0  \\
     \checkmark & & & & Market-1501 & 86.5 (+0.7) & 94.2 & 98.3 & 98.9\\
     \checkmark & \checkmark & & & Market-1501 & 86.5 (+0.7) & 94.2 & 98.3 & 98.9 \\
     \checkmark & \checkmark & \checkmark & & Market-1501 & 86.5 (+0.7) & 93.9 & 98.2 & 98.9\\
     \checkmark & \checkmark & \checkmark & \checkmark & Market-1501 & 86.9 (+1.1) & 94.4 & 98.3 & 99.0\\
     \bottomrule
    \end{tabular*}
    \vspace{+5mm}
    \caption{Component analysis. V: Variation-Normalized Autoencoder. J: Jensen-Shannon Triplet Loss. H:  Hierarchical Variation Distiller. C: Covariance  constraint. }\label{table:component-analysis}
    \vspace{-10mm}
    \end{center}

\end{table}
\begin{wraptable}{r}{0.35\linewidth}
\vspace{-7mm}
\centering
\fontsize{7}{8}\selectfont
  \begin{tabular}[b]{lcccc}
    \toprule
    J & $\gamma$ &  mAP &  Rank-1 & Rank-5 \\
    \hline\hline
    & - & 50.1 & 74.3 &85.9 \\
    \midrule
    \checkmark & 0.3 &  50.1 &  74.4 & 86.1 \\ 
    \checkmark & 0.4 &  50.3 &  74.2 & 86.2 \\
    \checkmark & \boldtext{0.5} &  \boldtext{50.6} &  \boldtext{75.0} & \boldtext{86.3} \\
    \checkmark & 0.6 &  50.2 & 74.3 & 86.1 \\
    \checkmark & 0.7 &  50.0 & 74.3 & 85.9 \\
    \bottomrule
    \end{tabular} 
\caption{\small Analyses on Jensen-Shannon triplet loss and margin value $\gamma$. J: JS triplet loss.}
\label{tab:jstriplet}
\end{wraptable}

\subsubsection{Variation-Normalized Autoencoder}
Firstly, we experiment on Variation-Normalized Autoencoder, which our later proposed methods developed from. In the experiment, the hyper-parameter $\alpha$ and $\beta$ are set to 0.2 and 1.0 for Market-1501 dataset, 0.2 and 2.0 for MSMT17 dataset. The result in Table~\ref{table:component-analysis} shows that VNAE outperforms the baseline model by an observable margin, especially on the larger datasets MSMT17 where the intra-class variations signify. The experiment results indicate that our variation normalization is effective, particularly on large-scale datasets.

\subsubsection{Jensen-Shannon Triplet Loss}
To optimize the training of VNAE, we replace the conventional triplet loss with Jensen Shannon triplet loss. We set the loss ratio $\lambda$ for JS triplet to 1.0. In Table \ref{tab:jstriplet}, we analyse margin $\gamma$'s effect to the performance of JS triplet loss. The experiments show that setting the $\gamma$ to 0.5 results in best performance on MSMT17 dataset, which improves 0.5 mAP from the previous VNAE's performance. The success of combining latent distribution learning and the Jensen-Shannon triplet loss not only supports the novel metric loss's potency, but also implies that our intra-class variation modelling is meaningful.

\subsubsection{Hierarchical Variation Distiller}
To enhance the representation quality, we employ Hierarchical Variation Distiller module to complete the HAVANA framework. The hyper-parameter $\alpha$ and $\beta$ are set to 0.2 and 1.0 for Market-1501 dataset, 0.2 and 2.0 for MSMT17 dataset. Table~\ref{table:component-analysis} shows a comparison before and after attaching the HVD module. The results show that a hierarchical representation can be beneficial for the Person Re-ID task.

\subsubsection{Covariance Constraint}
\begin{wrapfigure}{r}{0.45\textwidth}
\vspace{-15mm}
  \begin{center}
    \includegraphics[width=1.0\linewidth]{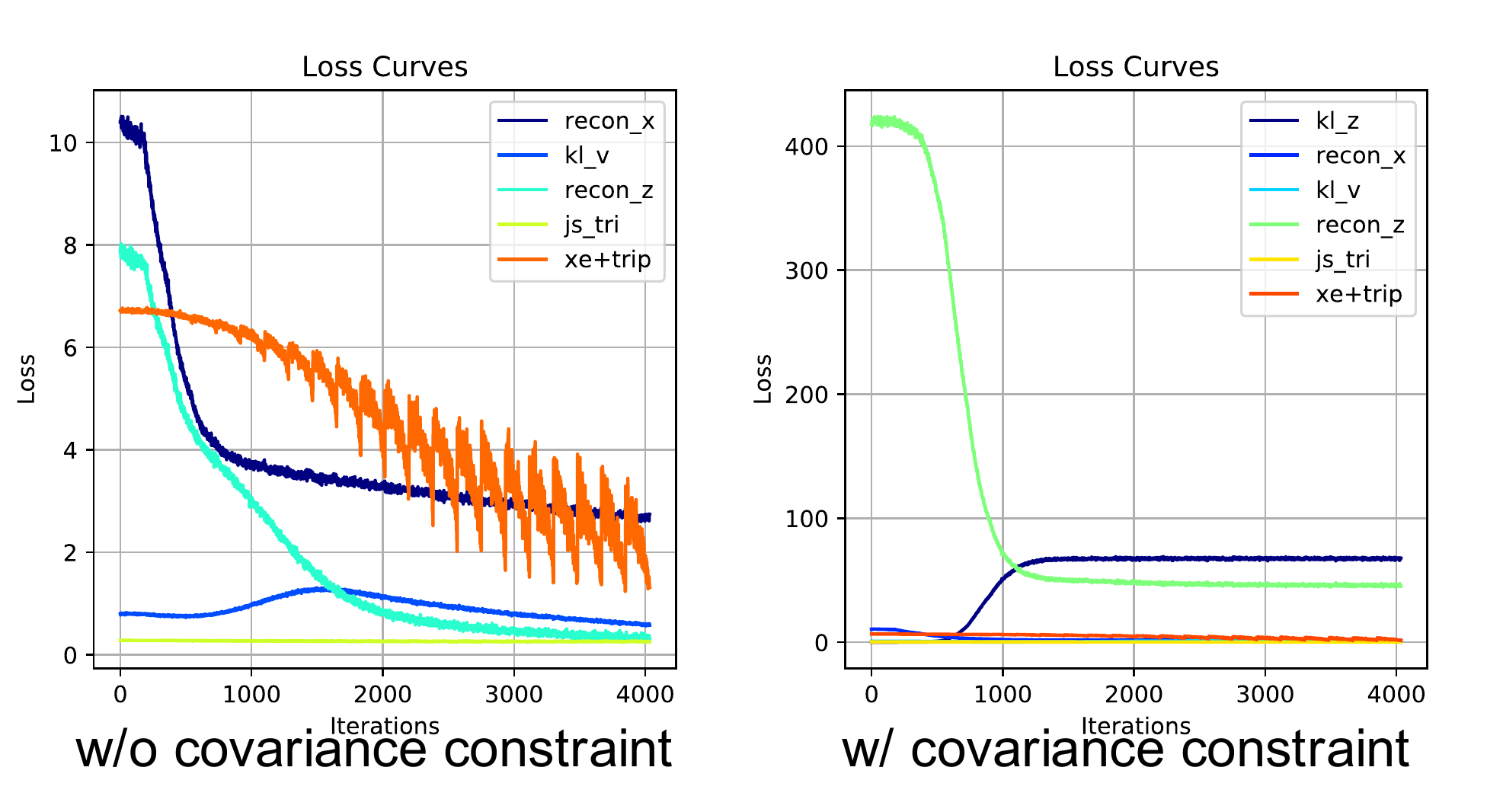}
  \end{center}
  \vspace{-5mm}
  \caption{\fontsize{9}{10}\selectfont Comparison of loss curves before and after imposing the covariance constraint (CC). Without CC, the loss curve of the generative loss $\mathcal{L}_\textrm{klv}$ dissipates in an early stage of training. With CC, the generative loss $\mathcal{L}_\textrm{klz}$ and the reconstruction loss $\mathcal{L}_\textrm{reconz}$ are maintained at a stable level.}
  \label{fig:cc}
  \vspace{-5mm}
\end{wrapfigure}
We put a covariance constraint on the latent representation by rewriting the original loss term 
$\mathcal{L}_\mathrm{klz}^\mathrm{ori} = \mathcal{L}_\mathrm{reconz} + \mathcal{L}_\mathrm{klz}$, where $\mathcal{L}_\mathrm{reconz} = \textrm{MSE}(\mathbf{z}_{\mu}, \tilde{\mathbf{z}}_{\mu})$ and $\mathcal{L}_\mathrm{klz}=D_\textrm{KL}(\mathcal{N}(\textbf{z}_\mu, \textbf{z}_\sigma^2) \parallel \mathcal{N}(\textbf{0}, \textbf{I}))$.
We show the comparison of training loss curves in Fig. \ref{fig:cc}. 
Without covariance constraint, the generative loss $L_{klv}$ dissipates due to a well-known phenomenon, posterior collapse~\cite{Lucas2019UnderstandingPC}, which leads the representation to be less meaningful. Covariance constraint balances the generative loss $L_{klz}$ and the reconstruction loss $L_{reconz}$ and improves representation learning.

Table \ref{table:component-analysis} shows that our proposed HAVANA performs better when trained with the covariance constraint, which not only justifies our intuition, but also suggests that the bias stemmed from unseen data can be alleviated in such a way.

\begin{table}[t]

\centering
\begin{tabular}{ ll cccc cccc}
\toprule
Method & \multicolumn{4}{c}{Market-1501} & \multicolumn{4}{c}{MSMT17} \\ \cmidrule{2-9}
       & \makebox[1cm][c]{mAP} & \makebox[1cm][c]{R-1} & \makebox[1cm][c]{R-5} & \makebox[1cm][c]{R-10} &  \makebox[1cm][c]{mAP} & \makebox[1cm][c]{R-1} & \makebox[1cm][c]{R-5} & \makebox[1cm][c]{R-10} & \\ \midrule

 baseline  & 85.8 & 94.2 & 98.4 & 99.0 & 48.2 & 72.2 & 83.9 & 87.7  \\
+$\mathcal{L}_\textrm{reconx}$ & 85.7 & 93.8 & 98.0 & 98.9 & 47.9 & 74.1 & 85.7 & 89.3 \\
+$\mathcal{L}_\textrm{klz}$& 86.5 & 94.2 & 98.3 & 98.9 & 50.1 & 74.3 & 85.9 & 89.6\\
 +$\mathcal{L}_\textrm{jst}$& 86.5 & 94.3 & 98.2 & 98.9 & 50.6 & 75.0 & 86.3 & 89.8 \\
+$\mathcal{L}_\textrm{reconz}$ & 86.6 & 94.5 & 98.2 & 99.0 & 52.3 & 76.5 & 87.2 & 90.4 \\
+$\mathcal{L}_\textrm{klv}$  & 86.9 & 94.4 & 98.3 & 99.0 & 52.3 & 76.7 & 87.2 & 90.6 \\
\bottomrule
\end{tabular}
\vspace{5mm}
\caption{Component analysis on all loss proposed in the HAVANA framework. We show that each loss contributes to final performance.}
\label{table:loss}
\vspace{0mm}
\end{table}
\subsection{Experiment Results}

\subsubsection{Overall Loss Analysis}
In this section, we use various experiments to analyse the contribution of each loss proposed in this paper, as shown in Table \ref{table:loss}. The analysis is conducted following the final scheme, HAVANA with covariance constraint. The results show that all losses proposed in the paper are important to the final performance.

\subsubsection{Analyses on Hyper-parameters $\alpha$ and $\beta$}
\begin{figure}[t] 
\hfill%
\begin{minipage}[]{0.55\textwidth} 
		\centering 
		\includegraphics[width=\textwidth]{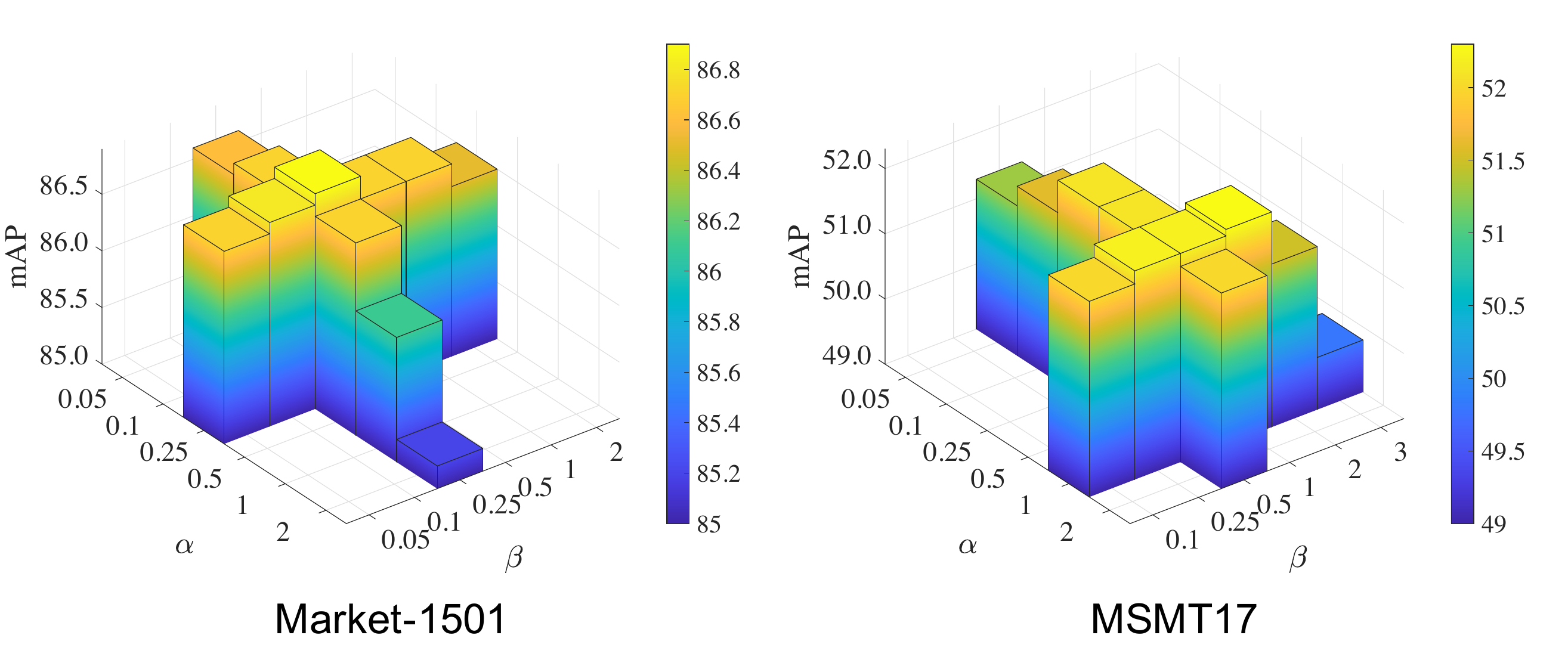} 
		\caption{3-D bar chart visualisation of hyper-parameter analysis. HAVANA is insensitive to $\alpha$ and $\beta$ values.}
		\label{fig:hyper-param}
	\end{minipage}%
\hfill%
	\begin{minipage}[]{0.4\textwidth}%
		\vspace{0pt}
		\centering
			
				\fontsize{7}{8}\selectfont
  \begin{tabular}[t]{lcccc}
     \toprule
    Dataset & Method & mAP & Rank-1\\
    \hline\hline
    Market-1501 & Baseline & 85.8 & 94.2 \\
    Market-1501 & Ours w/o RE & 86.2 & 94.3 \\
    Market-1501 & Ours w/ RE & 86.9 & 94.4  \\
    \hline\hline
    MSMT17 & Baseline & 48.2 & 72.2 \\
    MSMT17 & Ours w/o RE& 50.8 & 75.9 \\
    MSMT17 & Ours w/ RE & 52.3 & 76.7 \\
    \bottomrule
    \end{tabular} 
\tabcaption{Ablation study on artificial variation, Random Erasing.}
\label{table:random-erasing}
	\end{minipage}%
\hfill
\end{figure}

In Fig.\ref{fig:hyper-param}, we use 3-D bar chart to visualize hyper-parameters' effect on performance: two of the axis are $\alpha$ and $\beta$, and the other axis is the performance metric. The experiment results show that the optimal $\beta$ in a larger dataset, MSMT17, is higher than the optimal $\beta$ in a smaller dataset, Market-1501, which aligns with the fact that more intra-class variations need to be accommodated in a larger dataset.

\subsubsection{Effectiveness of Capturing Variations in Feature Space}
Random Erasing (RE) \cite{Zhong2017RandomED} is an augmentation method commonly seen in Person Re-ID works. RE randomly erases part of input images to imitate occlusions in real world. Our feature extractor is trained with the RE technique, which improves the extracted features' robustness to this artificial variation. To test whether HAVANA is capable of capturing such a subtle variation in the feature space, we train our model both with and without RE. The result in Table \ref{table:random-erasing} shows that our HAVANA performs better when with RE, which suggests that the artificial variation leaking through the feature extractor can be further filtered out by our model.


\subsubsection{Computational Cost}
\begin{wrapfigure}{r}{0.35\textwidth}
\vspace{-10mm}
  \begin{center}
    \includegraphics[width=1.0\linewidth]{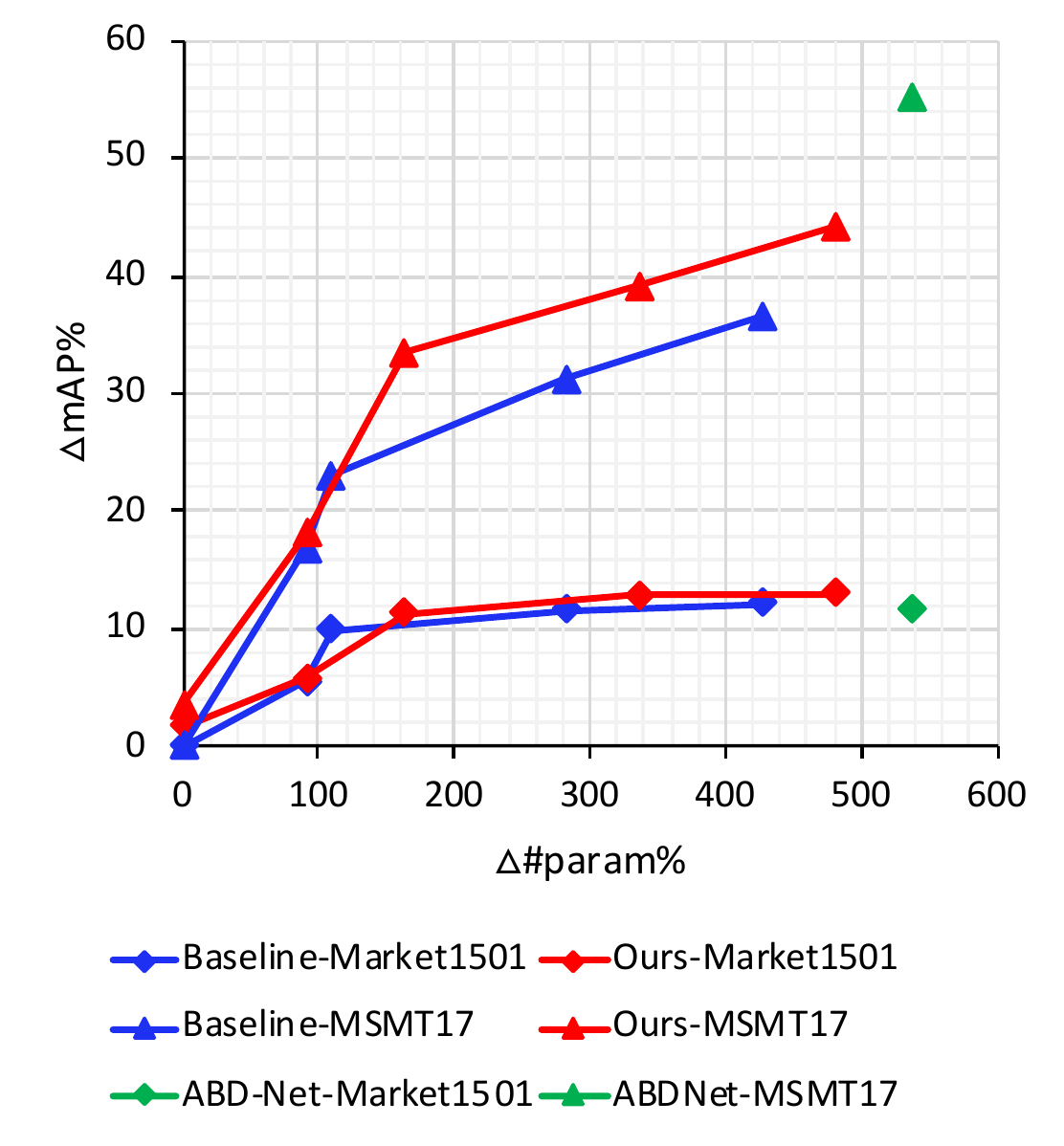}
  \end{center}
   \vspace{-5mm}
  \caption{\small Computational cost comparison. Our model (Red) shows consistent advantage over the baseline model (Blue) on the accuracy-parameter curve. Our model also  outperforms SOTA model (Green) with less parameters on Market1501. }
  \label{fig:param}
  \vspace{-5mm}
\end{wrapfigure}
In HAVANA, only the two-layer MLP that encodes from 2048-dimensional features to 2048-dimensional mean vectors is required test-time, whose parameters are $(2048+1) \times 2048 \times 2$. In Fig. \ref{fig:param}, we show a comparison of the performance increment against the number of parameters increment. 
We firstly experiment on baseline with various ResNet-like backbones.
In the figure, ResNet-18 baseline's result, which achieves 78.1 mAP on Market-1501 and 39.2 mAP on MSMT17 using 11M parameters, is set as the origin of the coordinate plane. 
We then experiment on HAVANA using baseline feature extractors previously obtained.
We compare with a SOTA method, ABD-Net~\cite{Chen_2019_ICCV_ABD-Net}, as well. The number of parameters in ABD-Net is recorded from the official implementation. Our method outperforms the baseline models consistently. HAVANA also beats the SOTA model with fewer parameters on Market-1501. We thus show that HAVANA is cost-efficient.



\section{Conclusion}

We present HAVANA, an extensible, light-weight hierarchical and variation-normalized autoencoder for the Person Re-ID task. We conducted extensive experiments and ablation studies on Market-1501, DukeMTMC-reID and MSMT17 datasets. We show that HAVANA significantly outperforms existing GAN-based methods on all datasets, and outperforms the SOTA Re-ID models on 2 datasets while achieving comparable results on the rest, with minimum supervision. We demonstrate VAE is naturally suitable to model variation in the Re-ID task.

However, our current HAVANA framework considers only basic VAE models. More complex VAE models, e.g., hyper-prior approaches~\cite{ansari2019hyperprior}, could be considered to improve disentangled representation learning.

\clearpage
%
%
\bibliographystyle{splncs04}
\bibliography{egbib}

\begin{thebibliography}{10}
\providecommand{\url}[1]{\texttt{#1}}
\providecommand{\urlprefix}{URL }
\providecommand{\doi}[1]{https://doi.org/#1}

\bibitem{Alemu_2019_ICCV}
Alemu, L.T., Pelillo, M., Shah, M.: Deep constrained dominant sets for person
  re-identification. In: The IEEE International Conference on Computer Vision
  (ICCV) (October 2019)

\bibitem{ansari2019hyperprior}
Ansari, A.F., Soh, H.: Hyperprior induced unsupervised disentanglement of
  latent representations. In: Proceedings of the AAAI Conference on Artificial
  Intelligence. vol.~33, pp. 3175--3182 (2019)

\bibitem{Arjovsky2017WassersteinG}
Arjovsky, M., Chintala, S., Bottou, L.: Wasserstein gan. ArXiv
  \textbf{abs/1701.07875} (2017)

\bibitem{BengioRepresentation}
{Bengio}, Y., {Courville}, A., {Vincent}, P.: Representation learning: A review
  and new perspectives. IEEE Transactions on Pattern Analysis and Machine
  Intelligence  \textbf{35}(8),  1798--1828 (Aug 2013).
  \doi{10.1109/TPAMI.2013.50}

\bibitem{Chen_2019_ICCV}
Chen, B., Deng, W., Hu, J.: Mixed high-order attention network for person
  re-identification. In: The IEEE International Conference on Computer Vision
  (ICCV) (October 2019)

\bibitem{Chen2018IsolatingSO}
Chen, T.Q., Li, X., Grosse, R.B., Duvenaud, D.: Isolating sources of
  disentanglement in variational autoencoders. In: NeurIPS (2018)

\bibitem{Chen_2019_ICCV_ABD-Net}
Chen, T., Ding, S., Xie, J., Yuan, Y., Chen, W., Yang, Y., Ren, Z., Wang, Z.:
  Abd-net: Attentive but diverse person re-identification. In: The IEEE
  International Conference on Computer Vision (ICCV) (October 2019)

\bibitem{Chen2016VariationalLA}
Chen, X., Kingma, D.P., Salimans, T., Duan, Y., Dhariwal, P., Schulman, J.,
  Sutskever, I., Abbeel, P.: Variational lossy autoencoder. ArXiv
  \textbf{abs/1611.02731} (2016)

\bibitem{Dai_2019_ICCV}
Dai, Z., Chen, M., Gu, X., Zhu, S., Tan, P.: Batch dropblock network for person
  re-identification and beyond. In: The IEEE International Conference on
  Computer Vision (ICCV) (October 2019)

\bibitem{Fang_2019_ICCV}
Fang, P., Zhou, J., Roy, S.K., Petersson, L., Harandi, M.: Bilinear attention
  networks for person retrieval. In: The IEEE International Conference on
  Computer Vision (ICCV) (October 2019)

\bibitem{NIPS2018_7398}
Ge, Y., Li, Z., Zhao, H., Yin, G., Yi, S., Wang, X., Li, h.: Fd-gan:
  Pose-guided feature distilling gan for robust person re-identification. In:
  Bengio, S., Wallach, H., Larochelle, H., Grauman, K., Cesa-Bianchi, N.,
  Garnett, R. (eds.) Advances in Neural Information Processing Systems 31, pp.
  1222--1233. Curran Associates, Inc. (2018),
  \url{http://papers.nips.cc/paper/7398-fd-gan-pose-guided-feature-distilling-gan-for-robust-person-re-identification.pdf}

\bibitem{Gong_2014_Springer}
Gong, S., Cristani, M., Yan, S., Loy, C.C.: Person Re-Identification (10 2014)

\bibitem{NIPS2018_7404}
Gonzalez-Garcia, A., van~de Weijer, J., Bengio, Y.: Image-to-image translation
  for cross-domain disentanglement. In: Bengio, S., Wallach, H., Larochelle,
  H., Grauman, K., Cesa-Bianchi, N., Garnett, R. (eds.) Advances in Neural
  Information Processing Systems 31, pp. 1287--1298. Curran Associates, Inc.
  (2018),
  \url{http://papers.nips.cc/paper/7404-image-to-image-translation-for-cross-domain-disentanglement.pdf}

\bibitem{NIPS2017_7159}
Gulrajani, I., Ahmed, F., Arjovsky, M., Dumoulin, V., Courville, A.C.: Improved
  training of wasserstein gans. In: Guyon, I., Luxburg, U.V., Bengio, S.,
  Wallach, H., Fergus, R., Vishwanathan, S., Garnett, R. (eds.) Advances in
  Neural Information Processing Systems 30, pp. 5767--5777. Curran Associates,
  Inc. (2017),
  \url{http://papers.nips.cc/paper/7159-improved-training-of-wasserstein-gans.pdf}

\bibitem{Gulrajani2016PixelVAEAL}
Gulrajani, I., Kumar, K., Ahmed, F., Ta{\"i}ga, A.A., Visin, F., V{\'a}zquez,
  D., Courville, A.C.: Pixelvae: A latent variable model for natural images.
  ArXiv  \textbf{abs/1611.05013} (2016)

\bibitem{Guo_2019_ICCV}
Guo, J., Yuan, Y., Huang, L., Zhang, C., Yao, J.G., Han, K.: Beyond human
  parts: Dual part-aligned representations for person re-identification. In:
  The IEEE International Conference on Computer Vision (ICCV) (October 2019)

\bibitem{He_2019_ICCV}
He, L., Wang, Y., Liu, W., Zhao, H., Sun, Z., Feng, J.: Foreground-aware
  pyramid reconstruction for alignment-free occluded person re-identification.
  In: The IEEE International Conference on Computer Vision (ICCV) (October
  2019)

\bibitem{higgins2017beta}
Higgins, I., Matthey, L., Pal, A., Burgess, C., Glorot, X., Botvinick, M.,
  Mohamed, S., Lerchner, A.: beta-vae: Learning basic visual concepts with a
  constrained variational framework. Iclr  \textbf{2}(5), ~6 (2017)

\bibitem{Hou_2019_CVPR}
Hou, R., Ma, B., Chang, H., Gu, X., Shan, S., Chen, X.:
  Interaction-and-aggregation network for person re-identification. In: The
  IEEE Conference on Computer Vision and Pattern Recognition (CVPR) (June 2019)

\bibitem{Kim2018DisentanglingBF}
Kim, H., Mnih, A.: Disentangling by factorising. In: ICML (2018)

\bibitem{Kingma2014SemisupervisedLW}
Kingma, D.P., Mohamed, S., Rezende, D.J., Welling, M.: Semi-supervised learning
  with deep generative models. ArXiv  \textbf{abs/1406.5298} (2014)

\bibitem{kingma2013auto}
Kingma, D.P., Welling, M.: Auto-encoding variational bayes. arXiv preprint
  arXiv:1312.6114  (2013)

\bibitem{NIPS2017_7178}
Lample, G., Zeghidour, N., Usunier, N., Bordes, A., DENOYER, L., Ranzato, M.A.:
  Fader networks:manipulating images by sliding attributes. In: Guyon, I.,
  Luxburg, U.V., Bengio, S., Wallach, H., Fergus, R., Vishwanathan, S.,
  Garnett, R. (eds.) Advances in Neural Information Processing Systems 30, pp.
  5967--5976. Curran Associates, Inc. (2017),
  \url{http://papers.nips.cc/paper/7178-fader-networksmanipulating-images-by-sliding-attributes.pdf}

\bibitem{Liu_2019_ICCV}
Liu, F., Zhang, L.: View confusion feature learning for person
  re-identification. In: The IEEE International Conference on Computer Vision
  (ICCV) (October 2019)

\bibitem{Liu_2018_CVPR}
Liu, J., Ni, B., Yan, Y., Zhou, P., Cheng, S., Hu, J.: Pose transferrable
  person re-identification. In: The IEEE Conference on Computer Vision and
  Pattern Recognition (CVPR) (June 2018)

\bibitem{Louizos2015TheVF}
Louizos, C., Swersky, K., Li, Y., Welling, M., Zemel, R.S.: The variational
  fair autoencoder. CoRR  \textbf{abs/1511.00830} (2015)

\bibitem{Lucas2019UnderstandingPC}
Lucas, J., Tucker, G., Grosse, R.B., Norouzi, M.: Understanding posterior
  collapse in generative latent variable models. In: DGS@ICLR (2019)

\bibitem{Luo_2019_ICCV}
Luo, C., Chen, Y., Wang, N., Zhang, Z.: Spectral feature transformation for
  person re-identification. In: The IEEE International Conference on Computer
  Vision (ICCV) (October 2019)

\bibitem{Luo_2019_Strong_TMM}
{Luo}, H., {Jiang}, W., {Gu}, Y., {Liu}, F., {Liao}, X., {Lai}, S., {Gu}, J.: A
  strong baseline and batch normalization neck for deep person
  re-identification. IEEE Transactions on Multimedia pp.~1--1 (2019).
  \doi{10.1109/TMM.2019.2958756}

\bibitem{Luo_2019_CVPR_Workshops}
Luo, H., Gu, Y., Liao, X., Lai, S., Jiang, W.: Bag of tricks and a strong
  baseline for deep person re-identification. In: The IEEE Conference on
  Computer Vision and Pattern Recognition (CVPR) Workshops (June 2019)

\bibitem{Miao_2019_ICCV}
Miao, J., Wu, Y., Liu, P., Ding, Y., Yang, Y.: Pose-guided feature alignment
  for occluded person re-identification. In: The IEEE International Conference
  on Computer Vision (ICCV) (October 2019)

\bibitem{Oord2017NeuralDR}
van~den Oord, A., Vinyals, O., Kavukcuoglu, K.: Neural discrete representation
  learning. In: NIPS (2017)

\bibitem{Qian_2018_ECCV}
Qian, X., Fu, Y., Xiang, T., Wang, W., Qiu, J., Wu, Y., Jiang, Y.G., Xue, X.:
  Pose-normalized image generation for person re-identification. In: The
  European Conference on Computer Vision (ECCV) (September 2018)

\bibitem{Quan_2019_ICCV}
Quan, R., Dong, X., Wu, Y., Zhu, L., Yang, Y.: Auto-reid: Searching for a
  part-aware convnet for person re-identification. In: The IEEE International
  Conference on Computer Vision (ICCV) (October 2019)

\bibitem{ergys2016performance}
Ristani, E., Solera, F., Zou, R.S., Cucchiara, R., Tomasi, C.: Performance
  measures and a data set for multi-target, multi-camera tracking. CoRR
  \textbf{abs/1609.01775} (2016)

\bibitem{Snderby2016LadderVA}
S{\o}nderby, C.K., Raiko, T., Maal{\o}e, L., S{\o}nderby, S.K., Winther, O.:
  Ladder variational autoencoders. In: NIPS (2016)

\bibitem{Sun_2019_CVPR}
Sun, Y., Xu, Q., Li, Y., Zhang, C., Li, Y., Wang, S., Sun, J.: Perceive where
  to focus: Learning visibility-aware part-level features for partial person
  re-identification. In: The IEEE Conference on Computer Vision and Pattern
  Recognition (CVPR) (June 2019)

\bibitem{Szegedy2015RethinkingTI}
Szegedy, C., Vanhoucke, V., Ioffe, S., Shlens, J., Wojna, Z.: Rethinking the
  inception architecture for computer vision. 2016 IEEE Conference on Computer
  Vision and Pattern Recognition (CVPR) pp. 2818--2826 (2015)

\bibitem{Tschannen2018RecentAI}
Tschannen, M., Bachem, O., Lucic, M.: Recent advances in autoencoder-based
  representation learning. ArXiv  \textbf{abs/1812.05069} (2018)

\bibitem{Wang2018LearningDF}
Wang, G., Yuan, Y., Chen, X., Li, J., Zhou, X.: Learning discriminative
  features with multiple granularities for person re-identification. In: MM '18
  (2018)

\bibitem{Wei_2018_CVPR}
Wei, L., Zhang, S., Gao, W., Tian, Q.: Person transfer gan to bridge domain gap
  for person re-identification. In: The IEEE Conference on Computer Vision and
  Pattern Recognition (CVPR) (June 2018)

\bibitem{Xia_2019_ICCV}
Xia, B.N., Gong, Y., Zhang, Y., Poellabauer, C.: Second-order non-local
  attention networks for person re-identification. In: The IEEE International
  Conference on Computer Vision (ICCV) (October 2019)

\bibitem{Yang_2019_CVPR}
Yang, W., Huang, H., Zhang, Z., Chen, X., Huang, K., Zhang, S.: Towards rich
  feature discovery with class activation maps augmentation for person
  re-identification. In: The IEEE Conference on Computer Vision and Pattern
  Recognition (CVPR) (June 2019)

\bibitem{Ye2020DeepLF}
Ye, M., Shen, J., jie Lin, G., Xiang, T., Shao, L., Hoi, S.C.H.: Deep learning
  for person re-identification: A survey and outlook. ArXiv
  \textbf{abs/2001.04193} (2020)

\bibitem{Yu_2019_ICCV}
Yu, T., Li, D., Yang, Y., Hospedales, T.M., Xiang, T.: Robust person
  re-identification by modelling feature uncertainty. In: The IEEE
  International Conference on Computer Vision (ICCV) (October 2019)

\bibitem{Zhang_2019_CVPR}
Zhang, Z., Lan, C., Zeng, W., Chen, Z.: Densely semantically aligned person
  re-identification. In: The IEEE Conference on Computer Vision and Pattern
  Recognition (CVPR) (June 2019)

\bibitem{zhao2017spindle}
Zhao, H., Tian, M., Sun, S., Shao, J., Yan, J., Yi, S., Wang, X., Tang, X.:
  Spindle net: Person re-identification with human body region guided feature
  decomposition and fusion. In: Proceedings of the IEEE conference on computer
  vision and pattern recognition. pp. 1077--1085 (2017)

\bibitem{Zhao2017InfoVAEIM}
Zhao, S., Song, J., Ermon, S.: Infovae: Information maximizing variational
  autoencoders. ArXiv  \textbf{abs/1706.02262} (2017)

\bibitem{Zheng2015ScalablePR}
Zheng, L., Shen, L., Tian, L., Wang, S., Wang, J., Tian, Q.: Scalable person
  re-identification: A benchmark. 2015 IEEE International Conference on
  Computer Vision (ICCV) pp. 1116--1124 (2015)

\bibitem{Zheng_2019_CVPR}
Zheng, Z., Yang, X., Yu, Z., Zheng, L., Yang, Y., Kautz, J.: Joint
  discriminative and generative learning for person re-identification. In: The
  IEEE Conference on Computer Vision and Pattern Recognition (CVPR) (June 2019)

\bibitem{Zheng_2017_ICCV}
Zheng, Z., Zheng, L., Yang, Y.: Unlabeled samples generated by gan improve the
  person re-identification baseline in vitro. In: The IEEE International
  Conference on Computer Vision (ICCV) (Oct 2017)

\bibitem{Zhong2017RandomED}
Zhong, Z., Zheng, L., Kang, G., Li, S., Yang, Y.: Random erasing data
  augmentation. ArXiv  \textbf{abs/1708.04896} (2017)

\bibitem{Zhong2017CameraSA}
Zhong, Z., Zheng, L., Zheng, Z., Li, S., Yang, Y.: Camera style adaptation for
  person re-identification. 2018 IEEE/CVF Conference on Computer Vision and
  Pattern Recognition pp. 5157--5166 (2017)

\bibitem{Zhou_2019_ICCV}
Zhou, K., Yang, Y., Cavallaro, A., Xiang, T.: Omni-scale feature learning for
  person re-identification. In: The IEEE International Conference on Computer
  Vision (ICCV) (October 2019)

\bibitem{Zhou_2019_ICCV_Discriminative}
Zhou, S., Wang, F., Huang, Z., Wang, J.: Discriminative feature learning with
  consistent attention regularization for person re-identification. In: The
  IEEE International Conference on Computer Vision (ICCV) (October 2019)

\end{thebibliography}
\end{document}